\documentclass[runningheads]{llncs}

\usepackage[T1]{fontenc}

\usepackage{graphicx}

\usepackage{hyperref}
\usepackage{color}

\usepackage{amssymb,algorithm2e}

\usepackage{longtable}

\usepackage{booktabs}
 
\usepackage{siunitx}

\usepackage[figuresright]{rotating}
\usepackage{tablefootnote}
\usepackage{enumitem}
\usepackage{floatrow}
\usepackage{subfigure}
\usepackage{bbding}

\author{Benedikt~Bollig\inst{1}\orcidID{0000-0003-0985-6115} \and
Matthias~F{\"u}gger\inst{1}\orcidID{0000-0001-5765-0301} \and
Thomas~Nowak\inst{1,2}\orcidID{0000-0003-1690-9342} \and 
Paul~Zeinaty\inst{1,3}\orcidID{0009-0003-2885-0410}\,\Envelope}

\authorrunning{B.~Bollig, M.~F{\"u}gger, T.~Nowak, and P.~Zeinaty}

\institute{Université Paris-Saclay, CNRS, ENS Paris-Saclay, LMF, Gif-sur-Yvette, France\\
\email{\{bollig,mfuegger,pzeinaty\}@lmf.cnrs.fr},
\email{thomas@thomasnowak.net}
\and
Institut Universitaire de France, Paris, France
\and
Direction Générale de l'Armement, Paris, France}

\usepackage{float}
\usepackage{array}
\usepackage{subcaption}
\usepackage{csvsimple}
\usepackage{colortbl}%
\usepackage{adjustbox}

\makeatletter
\def\FPS@rot{\rotatebox{90}}
\makeatother

\usepackage{tikz}
\usetikzlibrary{trees,positioning}
\usetikzlibrary{arrows.meta}
\usetikzlibrary{decorations.pathmorphing}

\usepackage{xspace}

\usepackage[most]{tcolorbox}
\usepackage{listings}
\usepackage{xcolor}
\definecolor{lightgray}{gray}{0.9}
\definecolor{codeblue}{rgb}{0.2,0.2,0.6}
\definecolor{codegreen}{rgb}{0,0.6,0}
\definecolor{codered}{rgb}{0.6,0,0}
\tcbset{colback=lightgray, colframe=lightgray, sharp corners, rounded corners, boxrule=0pt, left=5pt, right=5pt, top=-5pt, bottom=-5pt}
\newtcblisting{pythoncode}{
    listing only,
    listing options={
        language=Python,
        basicstyle=\ttfamily\small,
        keywordstyle=\color{codeblue},
        commentstyle=\color{codegreen},
        stringstyle=\color{codered},
        showspaces=false,
        showstringspaces=false,
    }
}

\newcommand{\eos}{\ensuremath{\mathsf{stop}}\xspace}

\newcommand{\Pred}{\ensuremath{p}\xspace}

\newcommand{\Act}{\ensuremath{\Sigma}\xspace}
\newcommand{\Acteos}{\ensuremath{{\Sigma_\eos}}\xspace}
\renewcommand{\epsilon}{\ensuremath{\varepsilon}\xspace}

\newcommand{\pref}{\mathit{pre}}

\newcommand{\figref}[1]{Figure \ref{#1}}

\begin{document}
\title{Promoting Simple Agents: Ensemble Methods for Event-Log Prediction}

\maketitle

\begin{abstract}
We compare lightweight automata-based models ($n$-grams) with neural architectures (LSTM, Transformer) for next-activity prediction in streaming event logs. Experiments on synthetic patterns and five real-world process mining datasets show that $n$-grams with appropriate context windows achieve comparable accuracy to neural models while requiring substantially fewer resources.
Unlike windowed neural architectures, which show unstable performance patterns, $n$-grams provide stable and consistent accuracy.
While we demonstrate that classical ensemble methods like voting improve $n$-gram performance, they require running many agents in parallel during inference, increasing memory consumption and latency.
We propose an ensemble method, the promotion algorithm, that dynamically selects between two active models during inference, reducing overhead compared to classical voting schemes.
On real-world datasets, these ensembles match or exceed the accuracy of non-windowed neural models with lower computational cost.
\end{abstract}

\section{Introduction}

Many modern information systems produce continuous event logs as time-ordered activities associated with concrete cases.
Examples include a patient admitted to a hospital, the issuing of a service ticket, or the expiration of a certificate.
The logs often contain information that allows one to answer central questions about the corresponding process.
The discipline of process mining \cite{Aalst16} studies the systematic extraction of models from these logs to describe and predict process behavior.
Thereby, classical approaches operate on offline data, \textit{e.g.}, evaluating and suggesting an optimized process.
By contrast, recent approaches also target streaming (or online) questions about the process while it is generating the event log, thereby enabling new types of application fields that were not feasible with offline methods \cite{Burattin22}.
One such streaming-enabled question is the prediction of the next activities from the event log before they happen.
Applications are in health monitoring of patients and infrastructure, improved response-times by predictive resource allocations, and targeted countermeasures to threats, among many others.

\paragraph{Activity prediction in discrete, case-based event logs.}
In this work, we focus on logs of finitely many discrete activities. An event is a (case ID, activity) pair. A case is a sequence of events with the same case ID, representing a single process instance such as a patient treatment, user session, or server operation. 
Examples for cases are subjects like patients, user sessions, or servers. An event log is a collection of events from multiple cases.
For such logs, we study the task of next-activity prediction: given a case's prefix, predict its most likely next activity, including a distinguished symbol \eos to mark the end of a sequence.
We restrict attention to predictors whose output for a case depends only on that case's history, that is, where cross-case features can be neglected.
This allows us to consider the projection onto the subsequence related to a case, individually.

Many such language models have been proposed in the literature, ranging from automata-based methods \cite{shannon1948mathematical,delaHiguera2010,Vaandrager17} to deep learning methods including recurrent architectures \cite{hochreiter1997long} and attention-based models \cite{vaswani2017attention}.
While such language models constantly increase in complexity, alternatives to monolithic language models via agent-based designs have shown promising results \cite{YaoZYDSN023,SchickDDRLHZCS23}, where specialized agents act as oracles called by a central, high-parameter LLM.
While such setups yield high accuracy, the complex LLM remains a bottleneck for throughput and latency, rendering this approach impractical for real-time processing of large event logs.

\paragraph{Stream prediction with simple language models.}
An interesting alternative to such resource-intensive LLMs are ensemble methods \cite{Zhou2012}, where the language model is obtained from simpler agents that operate in parallel on the log file, with their predictions being merged into a typically simple and fast aggregation function like a majority vote.
Similar strategies have been analyzed in the context of distributed computing to tolerate a certain fraction of misbehaving agents \cite{lamport1982byzantine,dolev1982efficient,dolev1982byzantine}, though typically motivated by hardware or software faults.

Inspired by these approaches, we study whether ensembles of simple language models that act in parallel perform as well as or outperform larger monolithic language models while achieving better throughput.

We propose a novel Promotion Algorithm that requires only 2 agents active in parallel during inference, substantially reducing overhead compared to classical ensemble methods like majority voting.

We evaluate this approach along with classical ensemble methods and baseline models on synthetic log streams that test specific properties such as counting and interleaving of events, as well as on 5 well-known real-life event-log datasets~\cite{sepsis_cases,bpi_2012,bpi_2013,bpi_2017,bpi_2018}, ranging from 15,214 to 2,514,266 events.
While long short-term memory (LSTM) networks \cite{hochreiter1997long} can outperform individual $n$-gram models, ensembles of simple $n$-gram agents match or exceed the accuracy of these neural models stably over a large range of window sizes, with the Promotion Algorithm having only little computational overhead over a simple $n$-gram.

All experiments are conducted within the logicsponge framework~\cite{logicsponge-processmining}, a Python library for next-activity prediction algorithms.

\paragraph{Related work.}
Our work touches upon several fields including automata learning and grammatical inference \cite{delaHiguera2010,Vaandrager17}, process mining \cite{Aalst16}, and various contexts within machine learning.

In automata learning, the Alergia algorithm \cite{carrasco1994learning} was proposed to learn stochastic regular languages from a set of words in a language.
The work in \cite{MaoCJNLN16} focuses on learning deterministic probabilistic automata from batches, building on variants of the Alergia algorithm.

Compared to classical automata learning in batch mode, grammatical inference for streaming data has received considerably less attention. Notable exceptions include \cite{BalleCG14,SchmidtK14,BaumgartnerV23}, which propose efficient algorithms for probabilistic automata to address computational complexities in streaming settings. Non-quantitative incremental automata learning was explored in \cite{Dupont96}.

Event-log prediction in process mining is an active area of research. Recent
works include \cite{PolatoSBL18,CeciLFCM14,PegoraroUGA21,BreukerMDB16,AalstSS11}.
Beyond generic sequence models, predictive monitoring work proposes architectures tailored to event-log data. Transformer-based approaches adapt self-attention to event logs; for example, ProcessTransformer uses event embeddings with positional encodings and is trained over all prefix lengths to predict the next activity, event time, and remaining time \cite{BukhshSD21}.
To complement sequence-only encodings, some approaches exploit graph structure: they either combine graph representations with Transformers \cite{ElyasiAaS24} or apply GNNs (including hybrids with recurrent models) on directly-follows graphs to better capture loops and control-flow dependencies \cite{RamaManeiroVL21,LischkaRS25,WangD25}.
Process mining in streaming settings has also been studied (\textit{e.g.}, \cite{BurattinSA14,Burattin22,KrawczykC18,ZelstDA18}).
Note that \cite{KrawczykC18} suggest using ensemble methods in the presence of noise,
though not resorting to $n$-grams.

The use of $n$-grams dates back to \cite{shannon1948mathematical} and has been widely applied to a multitude of problems including business processes \cite{BreukerMDB16}.

Apart from prediction, another interesting application domain of automata learning is verification \cite{MaoCJNLN16,MayrYCPV23,Vaandrager17,Leucker06}.
For example, \cite{MaoCJNLN16} focuses on learning deterministic probabilistic automata from batches, building on variants of the Alergia algorithm~\cite{carrasco1994learning}.
However, targeting reactive systems, the focus of this work is on infinite words rather than making predictions based on finite historical data.

Finally, several general-purpose automata learning libraries have been developed, including \cite{VerwerH17,MuskardinAPPT22,BolligKKLNP10,IsbernerHS15}.

\paragraph{Outline.}
We introduce general language models in Section~\ref{sec:model}. In Section~\ref{sec:window}, we study a central parameter for many streaming algorithms: the size of the window of historical events it has access to.
Section~\ref{sec:ensemble} introduces ensemble methods for streaming event-log prediction, including our novel promotion algorithm, and presents experimental results on synthetic and real-world datasets.

\section{Model for Next-Activity Prediction}
\label{sec:model}

We start with some basic notation.
We fix a nonempty finite set $\Act$ of activities, also referred to as actions.
The set is not necessarily known in advance to a language model, though an algorithm (\textit{e.g.}, a neural network) may require specifying an upper bound on its size (as part of the embedding dimension in case of the neural network).
Activities are referred to as $A, B$, \textit{etc.}
We write $\Act^\ast$ for the set of finite sequences of activities.

In particular, the set includes the empty sequence $\epsilon$. Given $L \subseteq \Sigma^\ast$, we let $L^\ast$ denote the set of all finite sequences obtained by concatenating zero or more words from $L$. Note that $L^\ast$ includes $\epsilon$.

We denote by \eos\ a distinguished symbol not in $\Act$ that marks the end of a sequence. 
The extended alphabet $\Acteos = \Act \cup \{\eos\}$ contains all activities plus the stop symbol.
In the following, we let $\sigma$ range over $\Acteos$.

Activities occur in the context of cases, where each case represents a single execution of the process. 
Case IDs are usually drawn from a countably infinite set (strings or natural numbers).
Since our prediction functions depend only on individual case histories, we abstract from case IDs and view an event log as a multiset over $(\Acteos)^*$, where each element is a complete case trace.
Every position in any word of this multiset corresponds to an event.
Note that in a streaming setting with interleaved cases, case IDs would be needed to separate concurrent executions; here, we assume traces are provided separately.

A language model can take several forms.
Examples are models based on $n$-grams, bags, probabilistic automata, and recurrent neural networks.
In essence, every such language model defines a \emph{probabilistic prediction function} or, simply, prediction function $\Pred \colon \Sigma^\ast \to \Delta(\Acteos)$, where
$\Delta(\Acteos)$ is the set of probability distributions over $\Acteos$.
Given a sequence $w \in \Act^\ast$, applying~$\Pred$ yields a probability distribution over the set of possible next activities, including \eos.
In analogy to conditional probabilities, one usually writes
$\Pred(\sigma \,|\, w)$ for the probability $\Pred(w)(\sigma)$.
The focus in this work is on predicting the most likely next activity, which is readily obtained from a probabilistic prediction function.

\paragraph{$n$-Grams.}

For $n \ge 1$, an $n$-gram model defines a prediction function $\Pred$ where the probability $\Pred(\sigma \,|\, w)$ depends only on the last $n{-}1$ activities of the sequence $w$. We call this suffix the \emph{history}. For sequences shorter than $n{-}1$ activities, the entire sequence serves as the history. For example, in a 3-gram model, the sequences $AB$ and $AAAB$ share the same suffix $AB$ of length $n{-}1 = 2$. The model thus assigns the same probability distribution over next activities in both cases: $\Pred(\sigma \,|\, AB) = \Pred(\sigma \,|\, AAAB)$ for all $\sigma \in \Acteos$. During training, these probabilities are estimated from the frequency distribution of activities that followed the history $AB$ in the training data.

In our implementation, $n$-grams are represented as probabilistic deterministic finite automata (PDFAs) where each state corresponds to a history (a suffix of length up to $n{-}1$). This representation enables efficient streaming operation: state transitions during inference require only updating the current automaton state, which implicitly encodes the relevant suffix of length up to $n-1$. Training consists of incrementing frequency counters associated with each state. These operations are computationally inexpensive, contributing to the low prediction and training times observed in our experiments.

\paragraph{Neural Networks on log sequences: LSTM and Transformer.}
Recurrent neural networks (RNNs) are a class of neural networks designed to process sequential data by maintaining a hidden state that captures information from previous time steps. Like $n$-grams, RNNs define a prediction function $\Pred \colon \Sigma^\ast \to \Delta(\Acteos)$, but compute it through learned transformations of sequential hidden states rather than explicit frequency counting.
Long Short-Term Memory (LSTM) networks \cite{hochreiter1997long} are a specific type of RNN that addresses the vanishing gradient problem, allowing them to capture long-term dependencies in sequences effectively. LSTMs achieve this through the use of memory cells and gating mechanisms that regulate the flow of information.
Thus, LSTMs are a natural candidate for predicting next activities in log streams.

Another frequently used architecture to handle long histories is the Transformer \cite{vaswani2017attention}.
They employ self-attention mechanisms that weigh the importance of different parts of the input sequence when computing $\Pred(\sigma \,|\, w)$, enabling them to capture global dependencies effectively.

\section{A Central Parameter: Impact of Window Size on Accuracy}\label{sec:window}

While $n$-grams and LSTMs are inherently suitable for streaming due to their sequential nature (with incremental state transitions), Transformer-based models require modifications to handle streaming data effectively.
Since Transformers process entire sequences at once, they require reprocessing the entire sequence every time a new token arrives, inducing arbitrarily long context sizes and rapidly increasing computational cost with sequence length.

A common solution is to use sliding windows to truncate the input sequence to a fixed length (also referred to as context size), similarly to $n$-grams.
Although this approach is close to padding techniques used in batch learning, it brings new challenges in streaming, such as determining the appropriate window size and learning positional embeddings that accurately reflect the temporal order of events within the window.
Intuitively, a too small window size will not allow the language model to incorporate old, but important actions into its prediction, resulting in a low overall accuracy.
Likewise, a too large window size is expected to lead to large parameter numbers that are not trained with the available data, and thus, again a degradation of accuracy.
Further, overly complex language models have large latencies and low throughput, invalidating their real-time use for event logs.

To better understand how state-of-the-art models behave with bounded windows in streaming settings, we evaluated the impact of different window sizes on their performance, comparing them to $n$-grams of similar context lengths.

Towards that goal, we designed synthetic benchmarks to exhibit specific properties and phenomena relevant to real-life event logs, as well as to highlight the window-size impact on the performance of different models.
For each tested pattern, we provide as reference the best theoretical accuracy achievable by any language model with window size~$n$, for varying values of~$n$.
We compare this to the accuracy of LSTM- and Transformer-based models with equal window sizes to assess how well they capture the underlying patterns in the data.
Finally, we validate our results on real-world datasets.

In the following, we describe the different datasets, defined over the alphabet $\Act = \{A, B\}$.

\subsection{Synthetic Dataset I: Deterministic periodic patterns}
A motif that is expected to be of central importance in many datasets, is counting.
Counting languages, including periodic patterns, have been studied extensively in the context of neural network learning capabilities \cite{WeissGY18}.
We start with the simplest such sequences, namely deterministic sequences, where the next activity is fully defined by the previous activities, except for the final $\eos$.
An ideal language model with a large enough window size can thus obtain an accuracy of asymptotically (within the sequence length) 100\%; limited only by wrong predictions of the final $\eos$.

For a set $L \subseteq \Sigma^\ast$, let $\pref(L)$ denote the set of prefixes of sequences in $L$, concatenated with a final $\eos$ symbol. We consider two types of synthetic periodic patterns:
\begin{itemize}[itemsep=0pt, topsep=1pt]
  \item \emph{6-step symmetric}, denoted as $AAABBB$: sequences from $\pref(\{A^3B^3\}^\ast)$. An example is $AAABBBA\eos$.

  \item \emph{5-step asymmetric}, denoted as $AAABB$: sequences from
  $\pref(\{A^3B^2\}^\ast)$. An example is $AAABBAAAB\eos$.
\end{itemize}

By design, an optimal model with a context length of 3 or above (\textit{e.g.}, a sufficiently trained $n$-gram model with $n \geq 4$), should be able to capture the entire pattern and make accurate predictions, while models with shorter context lengths achieve limited accuracy.

\subsection{Synthetic Dataset II: Randomized periodic patterns}
After evaluating deterministic periodic patterns, we consider event logs that exhibit randomized patterns, to assess the generalization capacity of the language model.
Naturally, for such patterns, optimal language models cannot asymptotically achieve 100\% accuracy.
This is immediately seen from a process that uniformly at random produces $A$ or $B$ until it stops.
An optimal model can achieve at most 50\% accuracy asymptotically.

The selected patterns are slightly more complex to incorporate effects of the history, repeating uniform random picks over a set of predefined sequences of activities, so-called dictionaries $\mathcal{D}$.
The sequences are then from the set $\pref(\mathcal{D}^\ast)$.
\begin{itemize}[itemsep=0pt, topsep=1pt]
  \item \emph{3-step full-discriminative}, denoted as $xx\bar{x}$ and defined by $\mathcal{D} = \{AAB, BBA\}$.
  Examples are $AABBBA\eos$ and $AABAABBBAAA\eos$.
  
  \item \emph{4-step half-discriminative}, denoted as $xAxB$ and defined by the dictionary
  \(\mathcal{D} = \{AAAB, BABB\}.\)
  An example is $BABBBABBAA\eos$.
  
  \item \emph{6-step tier-discriminative}, denoted as $xABxBA$ and defined by the dictionary
  $\mathcal{D} = \{AABABA, BABBBA\}$. For example: $BABBBAAABABA\eos$.
\end{itemize}

In each pattern, certain positions act as \emph{markers} that indicate which dictionary word was selected. The patterns combine two common characteristics of process data: periodic repetition of activities (loops) and the selection of different variants within these repetitions.

The \emph{full-discriminative pattern} $xx\bar{x}$ has all three positions differing between the dictionary words ($AAB$ vs.\ $BBA$). To model this pattern, it is necessary to determine which 3-step variant is repeated, as every position provides discriminative information.

The \emph{half-discriminative pattern} $xAxB$ contains markers only at positions 1 and 3 (both $A$ or both $B$), while positions 2 and 4 are fixed ($A$ and $B$, respectively). Here, models must identify which 4-step variant is active while separating discriminative positions from routine activities, \textit{i.e.}, activities that occur in all variants.

The \emph{tier-discriminative pattern} $xABxBA$ has markers at positions 1 and 4, with fixed activities in between. Identifying the active 6-step variant requires maintaining information across longer stretches of non-discriminative activities.

\subsection{Methods}
We evaluated the performance of standard language models for their dependency on window sizes.
This includes an LSTM, a Transformer, and an $n$-gram, each with access to historical events through sliding windows of different sizes.
While $n$-grams and fixed-context-length Transformers inherently require bounded histories, we capped the history for the LSTM to compare its accuracy to the other models.
We also ran experiments for non-windowed versions of the Transformer (with increasing context-size) and the LSTM.

\paragraph{General experimental setup.} All experiments were performed on an 11th Gen Intel Core i9-11900K architecture (3.50GHz, 8 cores, 24GB RAM) with an NVIDIA GeForce RTX 3090 (24GB) running Ubuntu (24.04.1) and Python (3.12.3).
For training and inference of the LSTM and Transformer models, we used PyTorch with CUDA.

All models are trained with Adam optimizer, cross-entropy loss, learning rate $10^{-3}$, batch size $8$, for up to $20$ epochs, using early stopping on validation accuracy with patience $3$ and gradient clipping. Each dataset was split into a training set (70\%), a validation set (15\%), and a test set (15\%).

All inputs use explicit one-hot encoding (no embedding). 
The LSTM model consists of two stacked LSTM layers (batch-first) with hidden size $128$, followed by a linear output layer mapping to vocabulary size.
The Transformer is a single-layer, single-head causal encoder with model dimension equal to the vocabulary size, using Rotary Positional Embeddings (RoPE), which simplify generation of the input sequence by appending and removing actions without changing the positional encoding for the remaining actions, and followed by a linear output layer mapping to vocabulary size.

\paragraph{Fixed context-window (prefix) training.}
In the windowed setting, each training sequence $x_1 \dots x_T \in (\Acteos)^\ast$ is expanded into all prefixes $x_1 \dots x_k \in \Act^\ast$, with $k<T$, each predicting a \emph{single} target token $x_{k+1} \in \Acteos$.
Prefixes are truncated to the last $k$ tokens and left-padded to a fixed window length $k$, inducing additional processing time for windowed LSTM and Transformer.
The model outputs logits for all prefix positions, but the loss is computed \emph{only at the final timestep}.
In contrast, the non-windowed setting uses full sequences, predicts all next tokens in parallel, and applies the loss at every non-padding position, allowing unbounded context.

\paragraph{Theoretical best accuracies.}
For each synthetic dataset and window size, we compute the theoretically best achievable accuracy by listing all possible subsequences of the given window size (with Python, and manually for short ones to check) and computing the accuracy of the optimal predictor. 
When this best achievable accuracy reaches the proportion of random activities in the pattern, it cannot increase further; we therefore fix this value for all larger window sizes.

\subsection{Results}

\begin{figure}[ht]
  \centering

  \subfigure[$AAABBB$ with 50,000 events]{\label{fig:binary_6step_symmetric_025}%
    \includegraphics[width=0.43\linewidth]{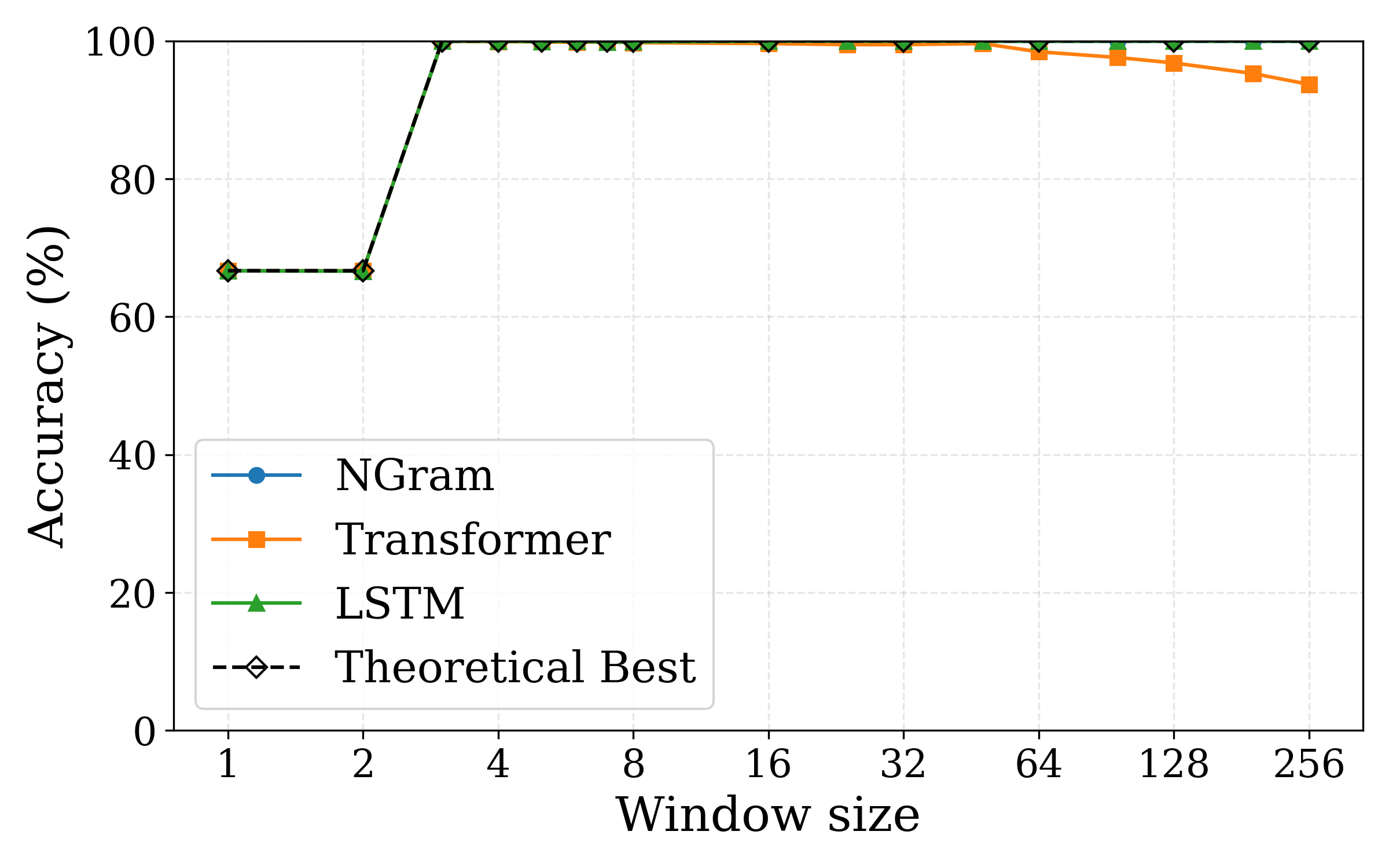}}
  \hfill
  \subfigure[$AAABBB$ with 200,000 events]{\label{fig:binary_6step_symmetric_1}%
    \includegraphics[width=0.43\linewidth]{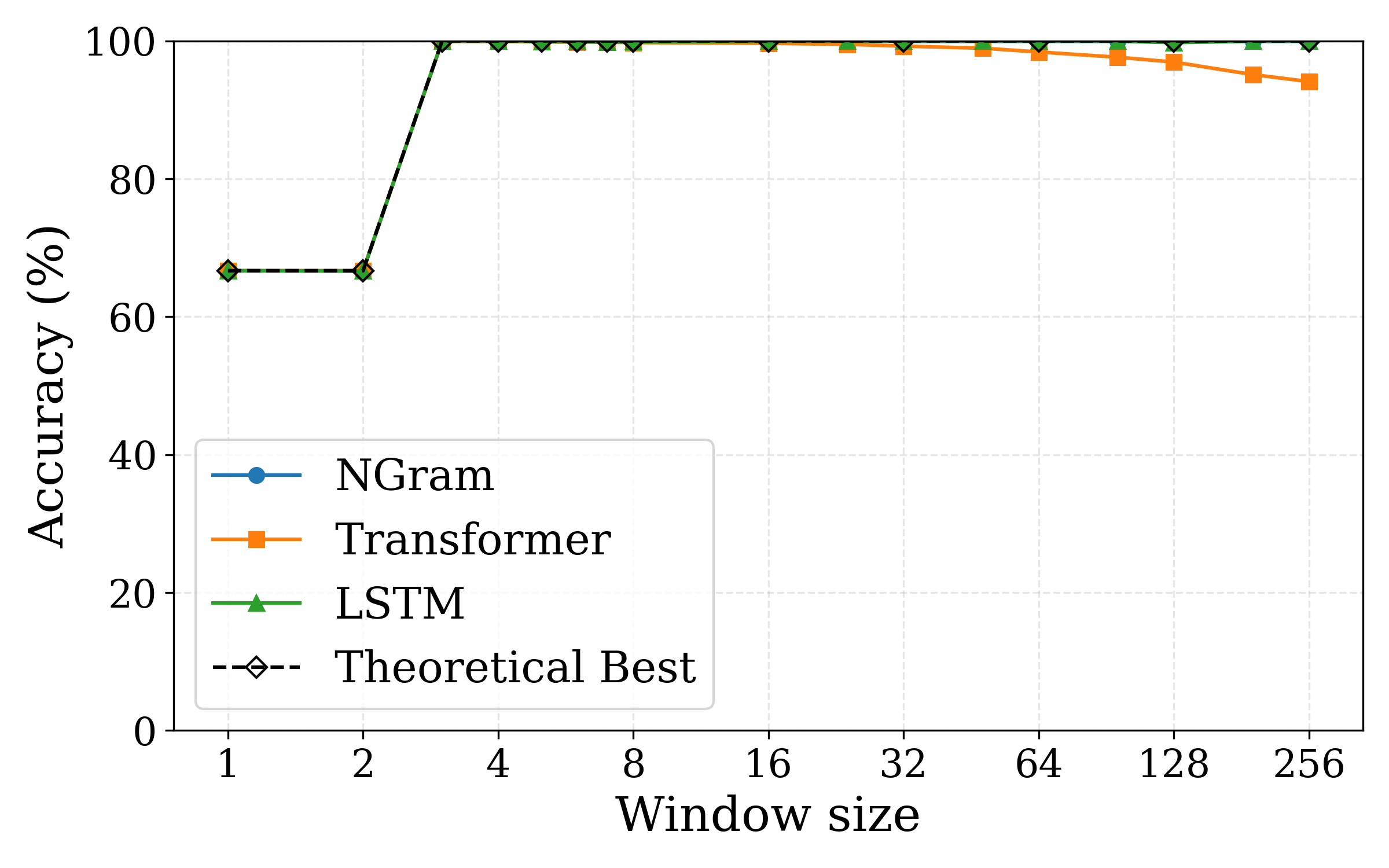}}

  \subfigure[$AAABB$ with 50,000 events]{\label{fig:binary_5step_asymmetric_025}%
    \includegraphics[width=0.43\linewidth]{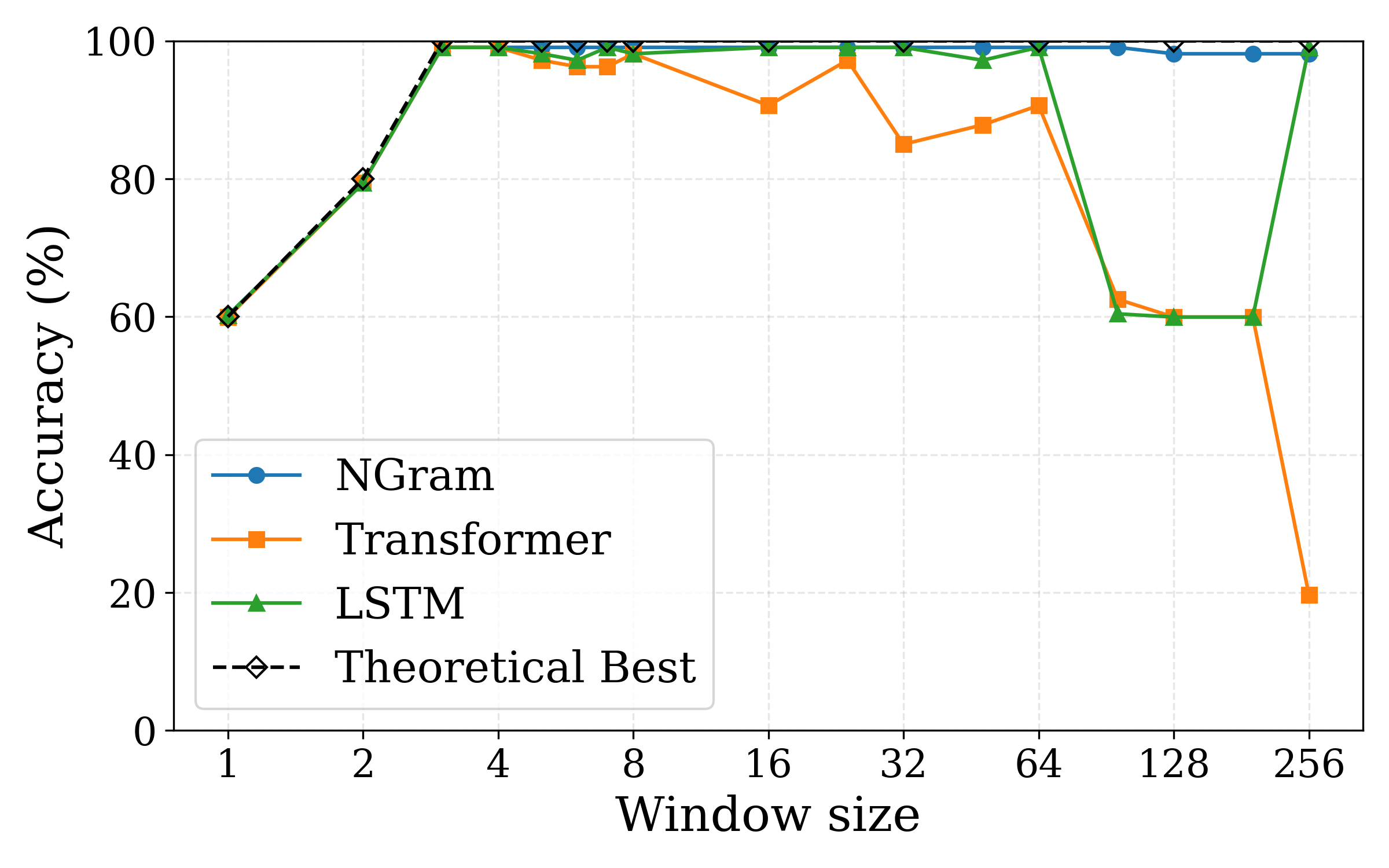}}
  \hfill
  \subfigure[$AAABB$ with 200,000 events]{\label{fig:binary_5step_asymmetric_1}%
    \includegraphics[width=0.43\linewidth]{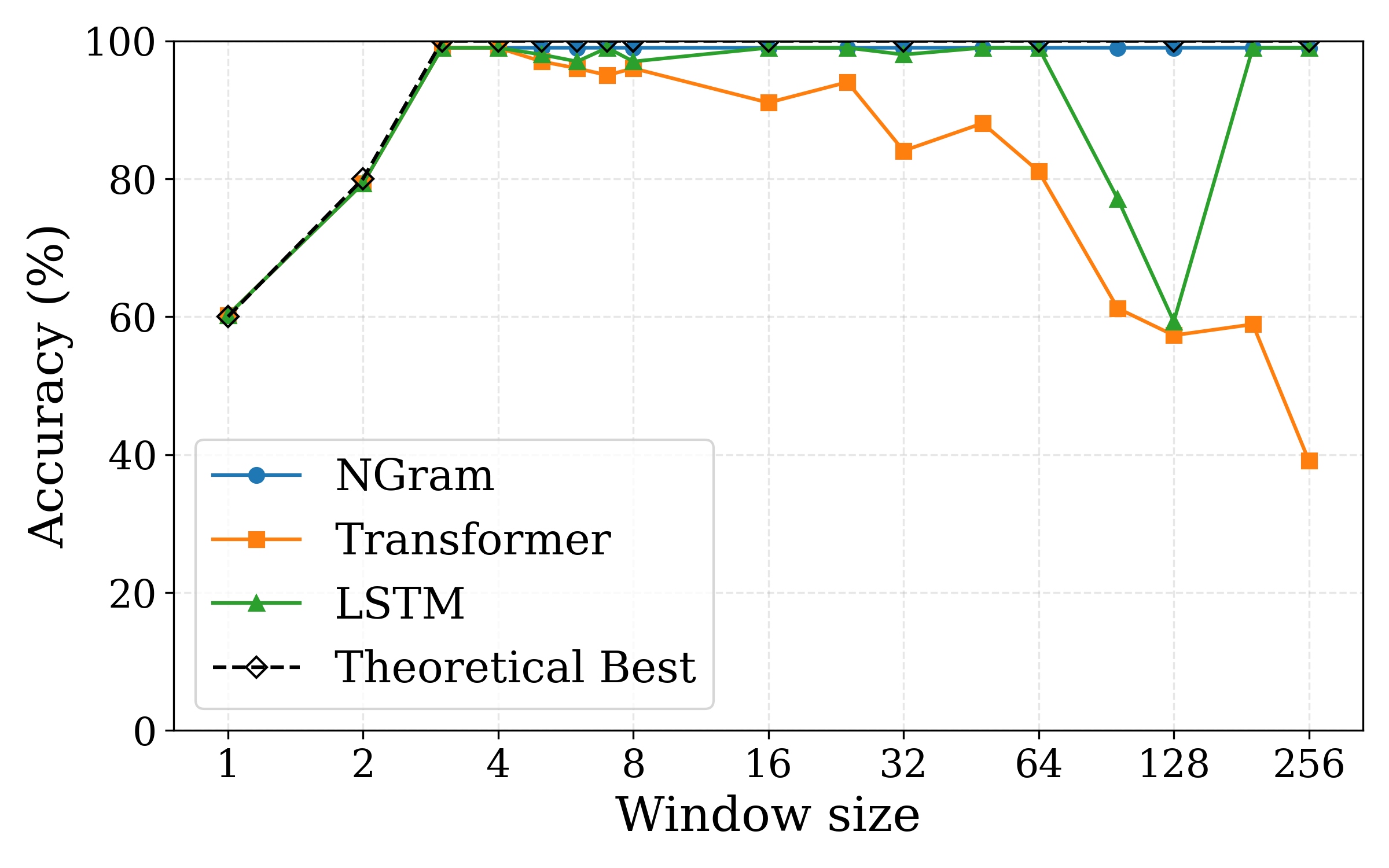}}
    
  \caption{Window-size impact on next-activity prediction accuracies for deterministic periodic patterns, with varying dataset sizes. The left side shows training on the first quarter and the right side training on the full dataset.}
  \label{fig:window_size_comparison}
\end{figure}

\begin{figure}[htb]
  \centering

  \subfigure[$xx\bar{x}$ with 50,000 events]{\label{fig:binary_3step_full_discriminative_025}%
    \includegraphics[width=0.40\linewidth]{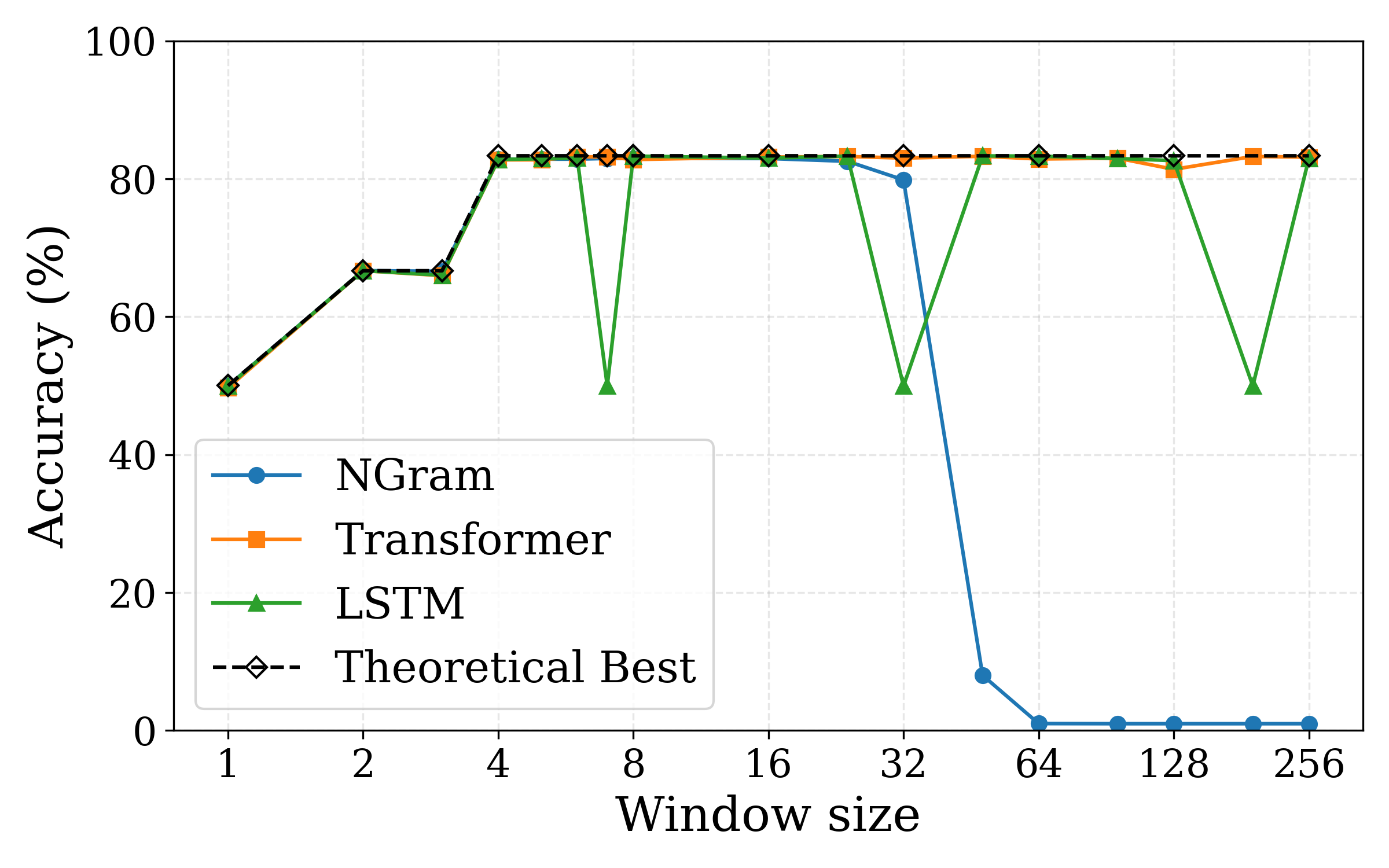}}
  \hfill
  \subfigure[$xx\bar{x}$ with 200,000 events]{\label{fig:binary_3step_full_discriminative_1}%
    \includegraphics[width=0.40\linewidth]{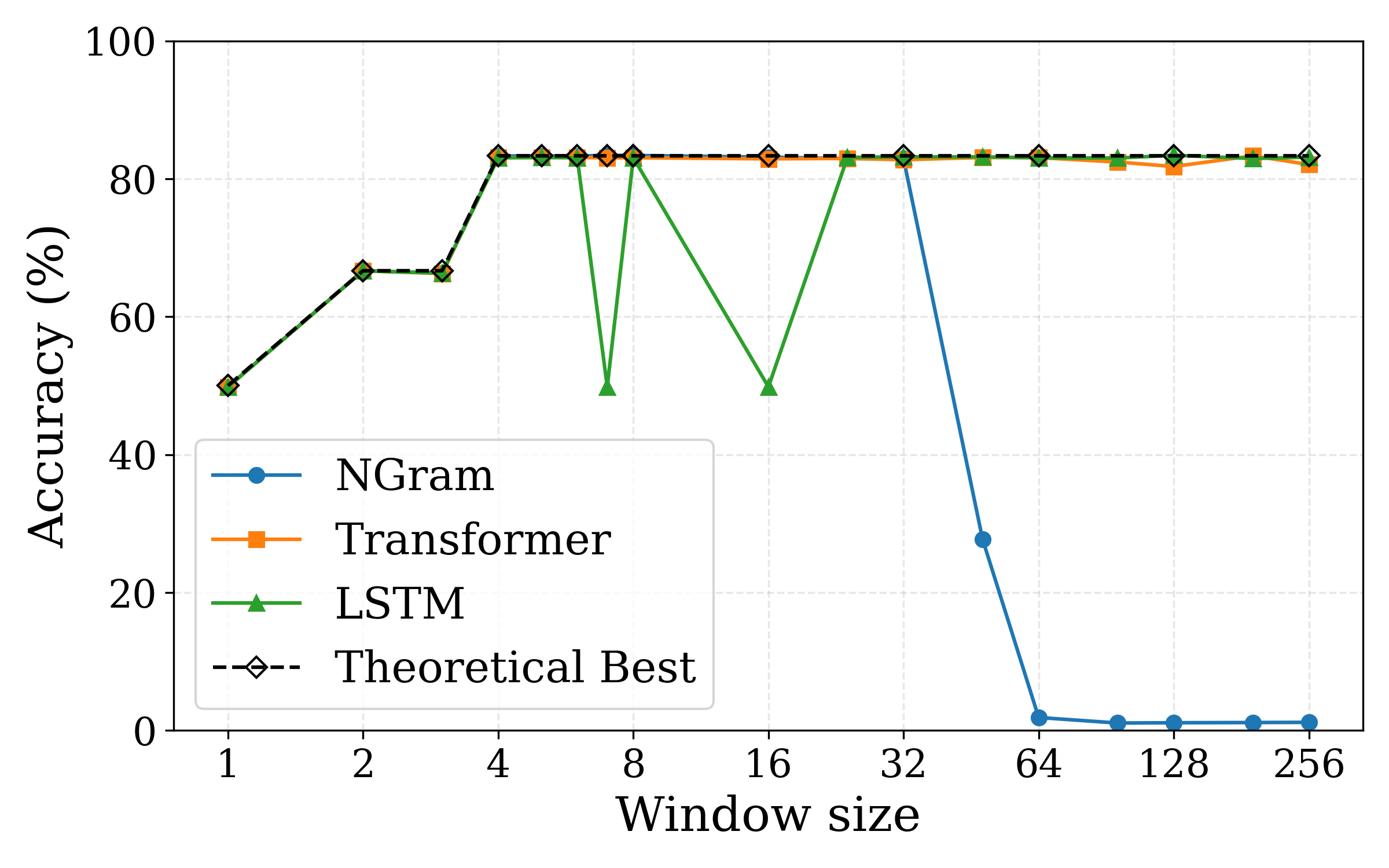}}

  \subfigure[$xAxB$ with 50,000 events]{\label{fig:binary_4step_half_discriminative_025}%
    \includegraphics[width=0.40\linewidth]{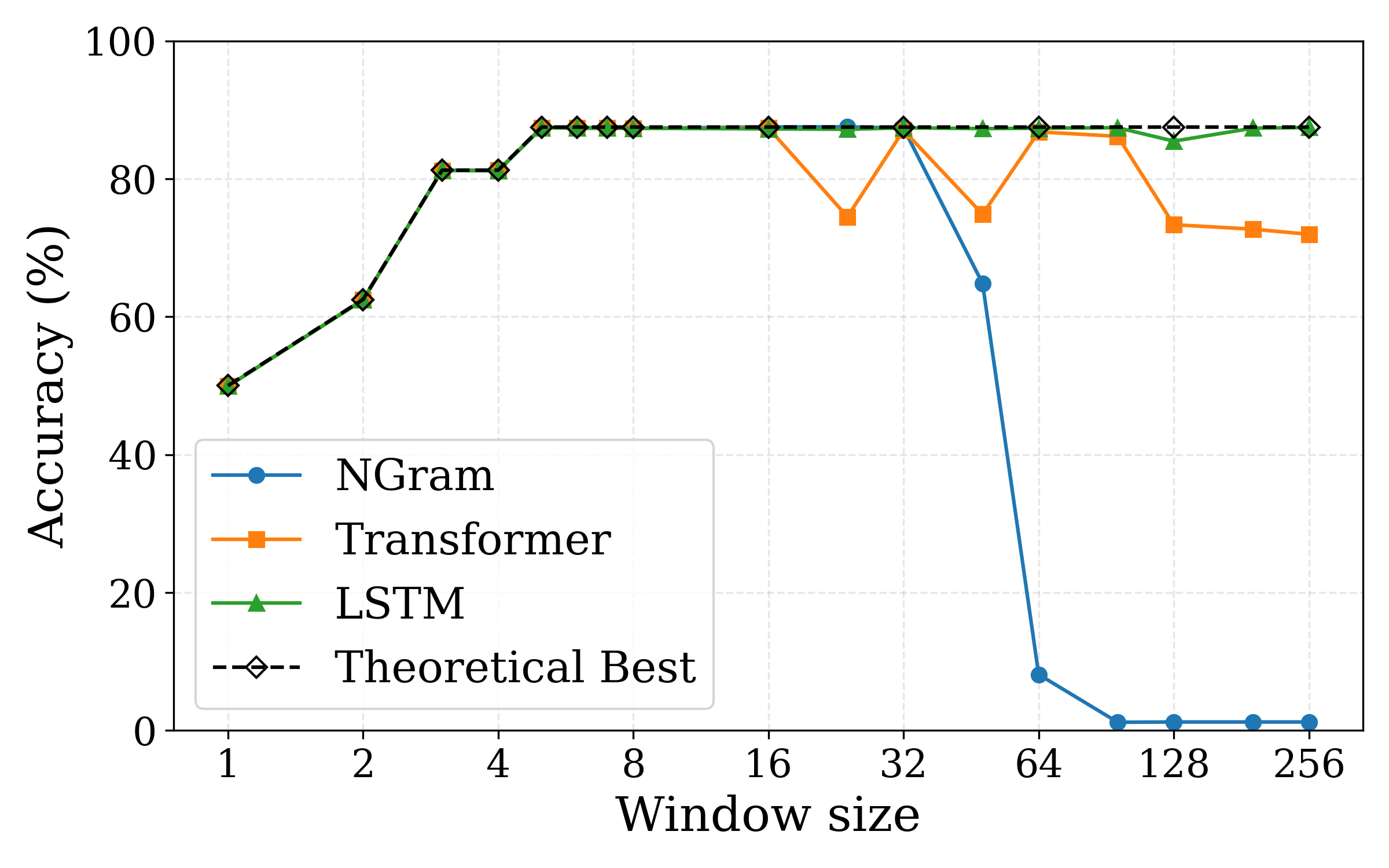}}
  \hfill
  \subfigure[$xAxB$ with 200,000 events]{\label{fig:binary_4step_half_discriminative_1}%
    \includegraphics[width=0.40\linewidth]{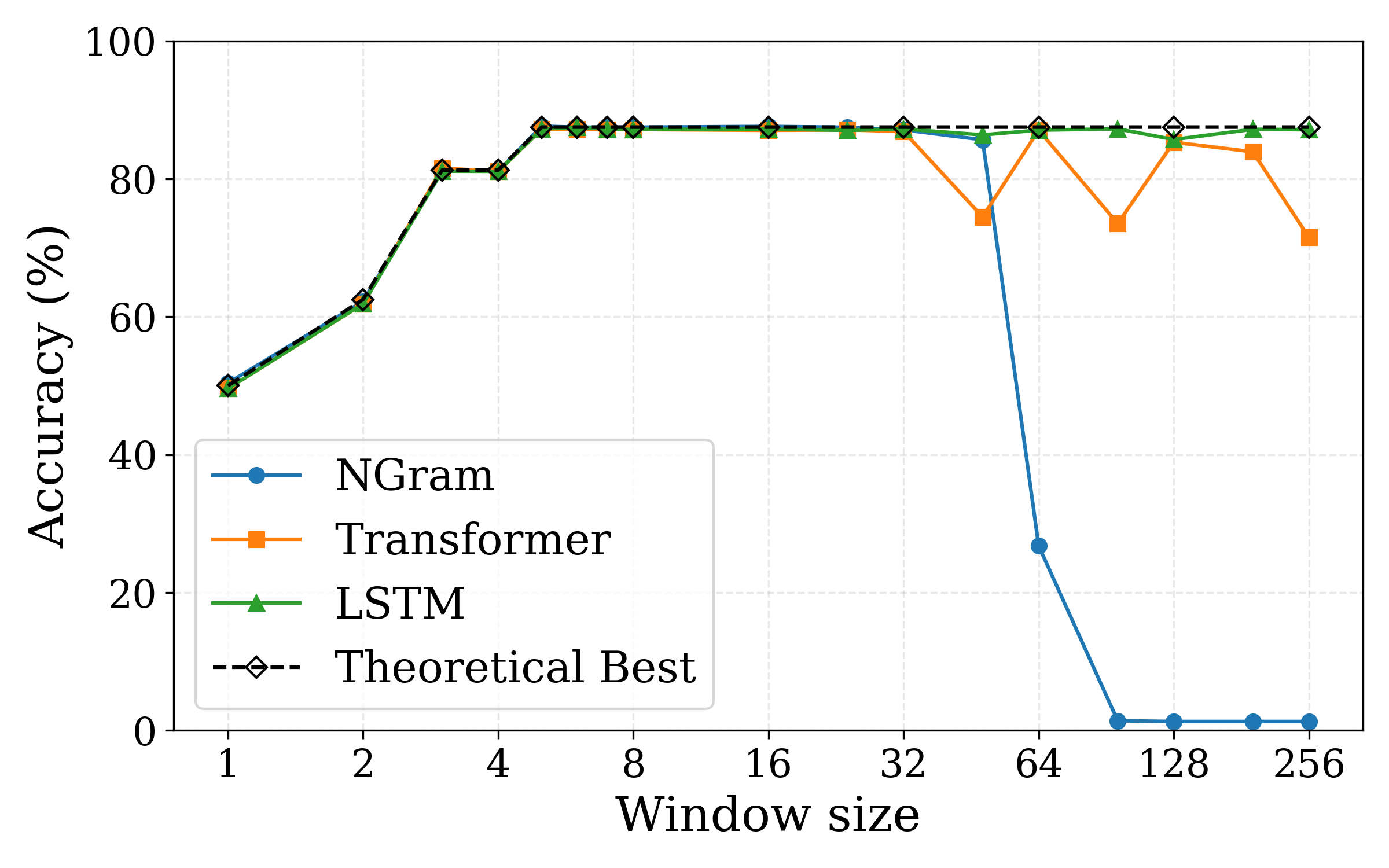}}

  \subfigure[$xABxBA$ with 50,000 events]{\label{fig:binary_6step_tier_discriminative_025}%
    \includegraphics[width=0.40\linewidth]{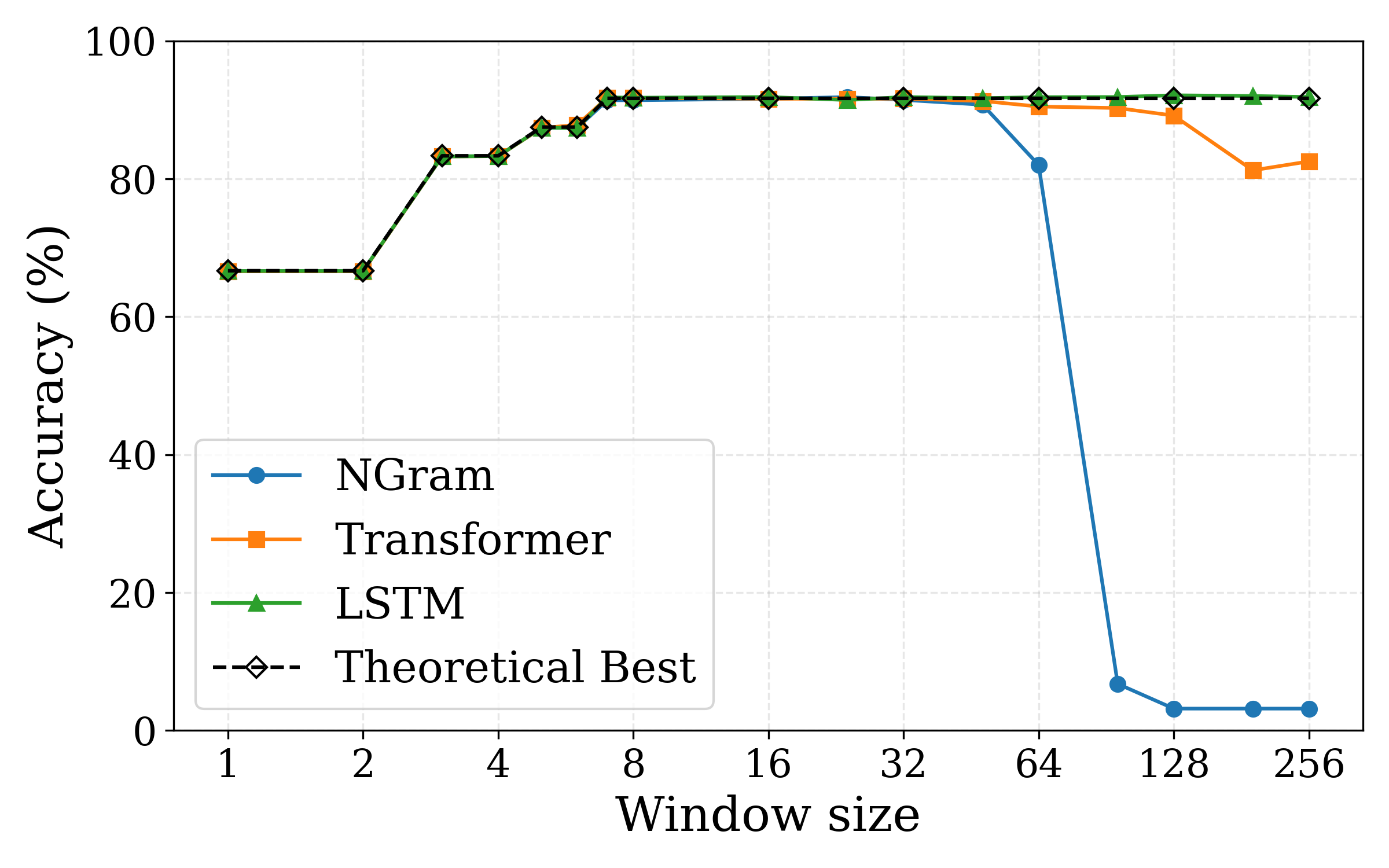}}
  \hfill
  \subfigure[$xABxBA$ with 200,000 events]{\label{fig:binary_6step_tier_discriminative_1}%
    \includegraphics[width=0.40\linewidth]{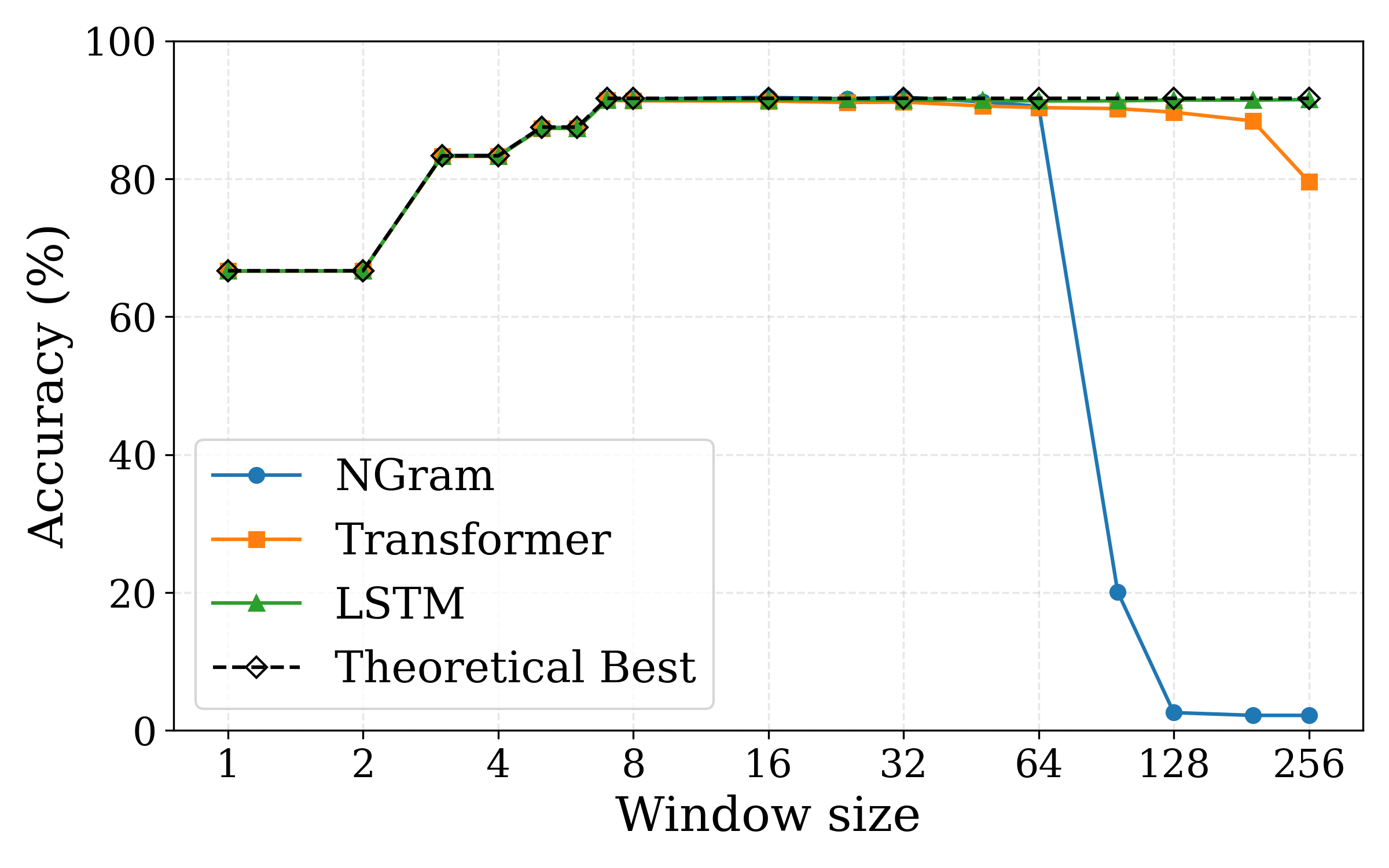}}

  \caption{Window-size impact on next-activity prediction accuracies for randomized periodic patterns, with varying dataset sizes. The left side shows training on the first quarter and the right side training on the full dataset.}
  \label{fig:randomized_patterns}
\end{figure}

Results of the experiments are shown in \figref{fig:window_size_comparison} for the Synthetic Dataset I (deterministic periodic) and in \figref{fig:randomized_patterns} for the Synthetic Dataset II (randomized periodic).
We evaluated model accuracies independently for each pattern.
Accuracies for the first quarter (25 cases, 50,000 events) and the full dataset (100 cases, 200,000 events) are shown to assess how the accuracy improves with the dataset size.

\paragraph{6-step symmetric pattern ($AAABBB$).}
For this simple pattern, all models match exactly the best theoretical accuracy, at least when window sizes remain under a certain range.
Interestingly, the Transformer's accuracy starts to degrade when the window size exceeds 16, whereas LSTM remains stable.
This specific behavior of the Transformer can be explained by its attention mechanism, which may struggle to keep focus on the relevant parts of the input sequence when the window size becomes too large compared to its attention capacity (dependent on single attention head(s)), whereas the LSTM's recurrent structure allows it to maintain a more consistent representation of the sequence.

\paragraph{5-step asymmetric pattern ($AAABB$).}
In order to slightly increase the difficulty of the task, we tested the asymmetric version of the previous pattern, thus requiring activity-specific model awareness.
For pattern $AAABB$, we first observe a more pronounced performance degradation for the Transformer.
This is likely due to the fact that the asymmetry in the pattern requires the model to capture more specific dependencies between activities, which appears more and more challenging for the Transformer as the window size grows beyond what is minimally required to capture the pattern.
Another interesting observation is that LSTMs show occasional accuracy drops at certain high window sizes, though these are less severe and systematic than the Transformer's degradation.
On the one hand, this indicates greater architectural robustness to increasing window sizes.

\paragraph{3-step full-discriminative pattern ($xx\bar{x}$).}
For the simple randomized pattern, $n$-grams exhibit a severe accuracy drop when the window size becomes too large. This is expected due to data sparsity: randomness in the sequences means that longer prefixes are each observed fewer times in training, leading to poor probability estimates and eventual accuracy collapse.
In accordance with this explanation, an improvement of the $n$-gram for the full dataset is observed -- shifting the observed decline of accuracy to the right from \figref{fig:binary_3step_full_discriminative_025} to \ref{fig:binary_3step_full_discriminative_1}.
The performance of LSTM and Transformer is relatively stable, with a slight performance degradation for the Transformer at very large window sizes, and exceptional learning failures (50\% accuracy) for the LSTM at window sizes 7 and 16.

\paragraph{4-step half-discriminative ($xAxB$) and 6-step tier-discriminative ($xABxBA$) patterns.} For these randomized patterns, as expected, the same statistical effect can be witnessed for $n$-grams with high window sizes.
Similarly, the decline is pushed to the right when comparing the quarter and full dataset.

LSTMs are again consistently accurate independently of the window size, whereas the Transformer's accuracy degrades as previously observed.
Interestingly, the window size at which this degradation starts appears to be about the same as for the $n$-grams.

Comparing the Transformer's accuracy across the three selected randomized periodic patterns, we observe a gradation in terms of slope and consistency of the accuracy degradation, as follows (by increasing consistency, and decreasing slope): half discriminative, then tier-discriminative, and finally full-discriminative.
This ordering may be linked to the availability of discriminative markers, and thus ``how difficult it is to choose the right position for the attention head of the Transformer''.

\section{Robustness via Ensemble Algorithms}
\label{sec:ensemble}

The previous section highlighted the impact of window sizes on the performance of state-of-the-art models for event-log prediction.
Over a large range of window sizes, $n$-grams demonstrate both good accuracy and computational efficiency due to their algorithmic simplicity.
Importantly, in contrast to the generally high accuracies of LSTM and Transformer, the experiments on synthetic datasets showed that $n$-grams are relatively robust with respect to window size: while performance does degrade at very large windows, $n$-grams maintain stable accuracy over a broad plateau, whereas LSTM and Transformer exhibit sudden performance breakdowns at specific window sizes within this range.

Selecting an appropriate window size for the $n$-gram within the plateau remains a challenge.
We next studied if this challenge can be addressed by running an ensemble of agents with different sizes and aggregating their decision. 

All base models in an ensemble are trained independently and in parallel on the same training data. The ensemble methods differ only in their evaluation strategy:

\paragraph{Soft voting.}
A natural choice for aggregation is combining probability distributions from multiple $n$-grams with different window sizes by averaging them and selecting the activity with highest average probability.
The overhead of this technique remains non-negligible, however, since all agents have to actively run on the stream.

\paragraph{Adaptive voting.}
This method tracks the accuracy of each model during evaluation, updating accuracy scores after each prediction, and selects the most accurate model from the ensemble for the next prediction.

As with the previous method, this technique requires running a potentially large set of agents with different window sizes in parallel.

\paragraph{Promotion algorithm.}
While the previous ensemble methods can increase robustness and avoid the window-size selection problem, they require running many agents in parallel, which comes with high computational overhead.
Moreover, the ensemble must cover a wide range of window sizes: our experiments showed that the optimal window size depends both on the inherent process structure (seen in the theoretical accuracy plateau in Figures~\ref{fig:window_size_comparison} and \ref{fig:randomized_patterns}) and on the amount of training data (seen in the accuracy decline beyond the plateau for large window sizes with limited data).

Crucially, we observed that $n$-grams shift this decline rightward as more training data becomes available.
This observation motivates an ensemble algorithm that requires only 2 agents running in parallel.
The idea of the Promotion Algorithm (Algorithm~\ref{alg:promotion}) is to predict with an $n$-gram while simultaneously tracking an $n'$-gram with $n' > n$. If the $n'$-gram achieves better accuracy than the $n$-gram for $\tau \geq 1$ predictions, the algorithm promotes the $n'$-gram to become the active model. 
The promotion threshold $\tau$ prevents premature promotion due to short-term performance fluctuations. Requiring $\tau$ consecutive confirmations adds robustness: if $M_i$ is truly better, $M_{i+1}$ is unlikely to accumulate enough wins to trigger promotion; if $M_{i+1}$ is superior, it will reliably do so. In our experiments, we set $\tau = 20$ based on manual empirical calibration during preliminary tests. Smaller values made the decision process unstable and more sensitive to data perturbations, whereas larger values reduced reactivity and delayed promotion, both leading to lower predictive accuracy.

\begin{algorithm}[ht]
\caption{Promotion algorithm with two active language models.}
\label{alg:promotion}
\footnotesize
\hrule\vspace{0.3em}
\textbf{Parameters:}
Base models $\mathcal{M} = (M_{1},\dots,M_{\ell})$ with window sizes \(s_1 < s_2 < \dots < s_\ell\); threshold $\tau \in \mathbb{N}$.
\vspace{0.3em}\hrule\vspace{0.3em}
$i \gets 1$ \tcp*[r]{start with smallest window size}
$c \gets 0$ \tcp*[r]{promotion counter}
$\mathit{Acc}(M_j) \gets 0$ for all $j \in \{1, \dots, \ell\}$\;

\While{true}{
    obtain predictions $\hat{a}_i$ from $M_{i}$ and $\hat{a}_{i+1}$ from $M_{i+1}$ (if $i < \ell$)\;
    
    \textbf{output} $\hat{a}_i$ as the current prediction \tcp*[r]{use active model for prediction}

    observe new incoming activity $a$\;
    
    update $\mathit{Acc}(M_{{i}})$ and $\mathit{Acc}(M_{i+1})$ by comparing $\hat{a}_i$ and $\hat{a}_{i+1}$ with $a$

    update internal states of $M_{{i}}$ and $M_{i+1}$ with activity $a$
    
    \BlankLine
    \uIf{$i < \ell$ \textbf{and} $\mathit{Acc}(M_{{i+1}}) > \mathit{Acc}(M_{i})$}{
        $c \gets c + 1$ \tcp*[r]{increment promotion counter}
        \uIf{$c \ge \tau$}{
            $i \gets i + 1$ \tcp*[r]{promote the next model}
            $c \gets 0$ \tcp*[r]{reset the promotion counter}
            reset accuracy counter for $M_{i}$ \tcp*[r]{for fair comparison between $M_i$ and $M_{i+1}$}
        }
    }
}
\end{algorithm}

\paragraph{Experimental validation.}
We tested all these ensemble algorithms together with the previously described monolithic language models on the synthetic datasets as well as real-world datasets; see Table~\ref{table:dataset-stats}.

\begin{table*}[ht]
\centering
\caption{Central statistics of the synthetic and real-world datasets used in the validation of next-activity prediction. Synthetic datasets I and II are split into a quarter and the full dataset. Real-world datasets are from \cite{sepsis_cases,bpi_2012,bpi_2013,bpi_2017,bpi_2018}.
}
\label{table:dataset-stats}
\scalebox{0.85}{
\begin{tabular}{@{}l
>{\hspace{5pt}}c<{\hspace{5pt}}
>{\hspace{5pt}}c<{\hspace{5pt}}
>{\hspace{5pt}}c<{\hspace{5pt}}
>{\hspace{5pt}}c<{\hspace{5pt}}
>{\hspace{5pt}}c<{\hspace{5pt}}
>{\hspace{5pt}}c<{\hspace{5pt}}
>{\hspace{5pt}}c
}
\toprule%
 & \shortstack[c]{Synthetic\\(Quarter)}
 & \shortstack[c]{Synthetic\\(Full)}
 & \shortstack[c]{Sepsis\\(2016)}
 & \shortstack[c]{BPI\\2012}
 & \shortstack[c]{BPI\\2013}
 
 & \shortstack[c]{BPI\\2017}
 & \shortstack[c]{BPI\\2018}
 
 \\\midrule
 \#Activities & 2 & 2 & 16 & 24 & 13 
 & 26 & 41 
 
 \\
 \#Cases & 25 & 100 & 1,050 & 13,087 & 7,554 
 & 31,509 & 43,809 
 
 \\
 Avg.\ case length & 500 & 2,000 & 14.49 & 20.04 & 8.68 
 & 38.16 & 57.39 
 
 \\
 \#Events & 50,000 & 200,000 & 15,214 & 262,200 & 65,533 
 & 1,202,267 & 2,514,266 
 
 \\
 \bottomrule
 \end{tabular}
 }
\end{table*}

\renewcommand{\thefootnote}{\fnsymbol{footnote}}

The results for accuracies, average prediction-times, and average training-times are summarized in 
Table~\ref{table:streaming-synthetic-results} for the synthetic datasets and 
Table~\ref{table:streaming-real-life-results} for the real-world datasets.
Window sizes for windowed versions of the LSTM and the Transformer are annotated, as well as $n_i$'s for the $n_i$-grams used in the ensembles.
For the ensemble methods, these sets of individual $n$-grams were selected by manual empirical calibration in preliminary experiments, retaining the configurations that yielded the best overall predictive performance.

Non-windowed versions of Transformers and LSTMs are seen to suffer from accuracy drops in the synthetic dataset, already.
Among the windowed language models (with a history length of 4), the 5-gram achieves high accuracy despite its simplicity, and is even surpassed by the Promotion algorithm for more complex synthetic patterns.
As expected, prediction and training times are low for the 5-gram, and good for the Promotion algorithm.

\begin{table*}[t]
\centering
\caption{Results on synthetic datasets. $\Delta t_\text{pred}$ and $\Delta t_\text{train}$ denote the average per-activity prediction- and training-time. Best values are in bold.
}
\label{table:streaming-synthetic-results}
\scalebox{0.80}{
\begin{tabular}{@{}l
>{\hspace{5pt}}c<{\hspace{5pt}}
>{\hspace{5pt}}c<{\hspace{5pt}}
>{\hspace{5pt}}c<{\hspace{5pt}}
>{\hspace{5pt}}c<{\hspace{5pt}}
>{\hspace{5pt}}c<{\hspace{5pt}}
>{\hspace{5pt}}c<{\hspace{5pt}}
>{\hspace{5pt}}c
}
\toprule%
 & \shortstack[c]{$A^3B^3$}
 & \shortstack[c]{$A^3B^2$}
 & \shortstack[c]{$xx\bar{x}$}
 & \shortstack[c]{$xAxB$}
 & \shortstack[c]{$xABxBA$}
 & \shortstack[c]{$\Delta t_\text{pred}$\\ ($\mu s$)}
 & \shortstack[c]{$\Delta t_\text{train}$\\ ($\mu s$)}
 \\\midrule
 5-gram & \bf 99.95 & \bf 99.95 & \bf 83.31 & 81.11 & \bf 83.40
 & 
 24 & \bf 4

 \\
 LSTM[win:4] & \bf 99.95 & \bf 99.95 & 83.04 & 81.08 & 83.27
 &
932 & 2133

 \\
 Transformer[win:4] & \bf 99.95 & \bf 99.00 & 83.09 & 81.09 & 83.27
 &
2647 & 5213

 \\
 LSTM & 50.02 & 59.89 & 50.11 & 49.68 & 50.29
 &
\bf 11 & 43

 \\
 Transformer & 66.67 & 62.22 & 50.08 & 49.59 & 66.62
 &
20 & 43
 
 \\
 Soft voting {\footnotesize(3,4,5,6)} & \bf 99.95 & \bf 99.95 & 83.12 & 87.23 & 83.32
 &
32\footnotemark[1] & 9\footnotemark[1]

 \\
 Adaptive voting {\footnotesize(3,4,5,6)} & \bf 99.95 & \bf 99.95 & 83.11 &  87.23 & 83.32
 &
43\footnotemark[1] & 5\footnotemark[1]

 \\
 Promotion {\footnotesize(3,5,7,9,13,17,25,33)} & 99.82 & 99.88 & 83.27 & \bf 87.33 & 83.35
 &
26 & 10\footnotemark[1]
 
 \\
 \bottomrule
 \end{tabular}
 }
\end{table*}

\begin{table*}[t]
\centering
\caption{Results on real-world datasets. $\Delta t_\text{pred}$ and $\Delta t_\text{train}$ denote the average per-activity prediction- and training-time. Best values are in bold.}
\label{table:streaming-real-life-results}
\scalebox{0.80}{
\begin{tabular}{@{}l
>{\hspace{5pt}}c<{\hspace{5pt}}
>{\hspace{5pt}}c<{\hspace{5pt}}
>{\hspace{5pt}}c<{\hspace{5pt}}
>{\hspace{5pt}}c<{\hspace{5pt}}
>{\hspace{5pt}}c<{\hspace{5pt}}
>{\hspace{5pt}}c<{\hspace{5pt}}
>{\hspace{5pt}}c
}
\toprule%
 
 & \shortstack[c]{Sepsis\\(2016)}
 & \shortstack[c]{BPI\\2012}
 & \shortstack[c]{BPI\\2013}
 
 & \shortstack[c]{BPI\\2017}
 & \shortstack[c]{BPI\\2018}
 & \shortstack[c]{$\Delta t_\text{pred}$\\ ($\mu s$)}
 & \shortstack[c]{$\Delta t_\text{train}$\\ ($\mu s$)}
 \\\midrule
 5-gram & 62.46 & 84.83 & 72.61 & 86.99 & 71.90
 & 
 \bf 57
 & 
 \bf 6
 
 \\
 LSTM[win:4] &  51.50 & 73.35 & 63.27 & 81.38 & 69.51
 &
1106 & 3370

 \\
 Transformer[win:4] & 57.24 & 73.47 & 56.87 & 81.07 & 68.68
 &
 2696 & 8308
 
 \\
 LSTM & 64.91 & \bf 85.87 & \bf 73.89 & \bf 88.42 & \bf 82.07
 &
123 & 556
 
 \\
 Transformer & 62.86 & 85.78 & 73.69 & 88.40 & 79.91
 &
196 & 781
 
 \\
 Soft voting {\footnotesize(3,4,5,6)} & \bf 65.68 & 85.31 & 72.54 & 87.16 & 74.57
 &
97\footnotemark[1]
 &
  9\footnotemark[1]

 \\ 
 Adaptive voting {\footnotesize(3,4,5,6)} & 62.55 & 85.58 & 72.66 & 86.90 & 72.71
 &
 129\footnotemark[1]
 &
 9\footnotemark[1]

 \\
 Promotion {\footnotesize(3,5,7,9,13,17,25,33)} & 61.60 & 85.50 & 72.40 & 87.39 & 75.80
 & 62 & 50\footnotemark[1]
 
 \\
 \bottomrule

 \end{tabular}
 }
\end{table*}
\footnotetext[0]{\footnotesize \footnotemark[1]: Latencies for ensemble models include parallelization of the (active) base models.}

For the real-world data, LSTMs achieve the highest accuracy.
This is in contrast to the synthetic dataset.
However, the 5-gram is not far below, demonstrating again a consistently high accuracy across a wide range of datasets.
It is even surpassed by ensemble methods, in particular soft voting.
Again, the Promotion algorithm achieves comparable accuracy, without requiring the choice of a window size as a hyperparameter, and is faster in prediction than more complex ensemble methods.

\section{Conclusion}

Ensembles of agents have been assessed for their performance when compared to classical monolithic language models in the context of predicting next activities in log streams.
A particularly promising algorithm, the Promotion algorithm, has been identified for obtaining high accuracy at low total prediction and training latency.
Future work will be devoted to online learning of such algorithms, with emphasis on (i) automatic calibration of the promotion threshold $\tau$ and automatic selection of the $n$-gram sets used by ensemble methods, (ii) extending the Promotion algorithm with a demotion mechanism to improve adaptability under concept drift, and (iii) improving the promotion procedure to reuse the learned structure and parameters of lower-order $n$-grams to initialize higher-order models.

\begin{credits}
\subsubsection{\ackname} 
The work was supported by the French National Research Agency (ANR) projects DREAMY (ANR-21-CE48-0003) and COSTXPRESS (ANR-23-CE45-0013).
It was also supported by the SAIF project, funded by the ``France 2030'' government investment plan managed by ANR, under the reference ANR-23-PEIA-0006.
We thank the anonymous reviewers for their thoughtful and thorough feedback.

\subsubsection{\discintname}
The authors have no competing interests to declare that are
relevant to the content of this article.
\end{credits}

\bibliographystyle{splncs04}
\bibliography{lit}

\end{document}